\def\year{2019}\relax
\documentclass[letterpaper]{article} 
\usepackage{aaai19}  
\usepackage{times}  
\usepackage{helvet}  
\usepackage{courier}  
\usepackage{url}  
\usepackage{graphicx}  

\usepackage{amsmath}
\usepackage{amsfonts}
\usepackage{multirow,booktabs}
\usepackage{paralist}
\usepackage{latexsym}
\usepackage{amssymb,verbatim}

\begin{document}
%
\title{Learning Multi-Level Information for Dialogue Response Selection \\by Highway Recurrent Transformer}
\author{Ting-Rui Chiang\quad Chao-Wei Huang\quad Shang-Yu Su\quad Yun-Nung Chen \\
National Taiwan University, Taiwan \\
\texttt{\{r07922052,r07922069,f05921117\}@csie.ntu.edu.tw\quad y.v.chen@ieee.org}}
\maketitle

\begin{abstract}
    With the increasing research interest in dialogue response generation, there is an emerging branch formulating this task as selecting next sentences, where given the partial dialogue contexts, the goal is to determine the most probable next sentence.
    Following the recent success of the Transformer model \cite{vaswani2017attention}, this paper proposes (1) a new variant of attention mechanism based on multi-head attention, called \emph{highway attention}, and (2) a recurrent model based on transformer and the proposed highway attention, so-called \emph{Highway Recurrent Transformer}.
    Experiments on the response selection task in the seventh Dialog System Technology Challenge (DSTC7) show the capability of the proposed model of modeling both utterance-level and dialogue-level information; the effectiveness of each module is further analyzed as well.
\end{abstract}

\section{Introduction}

With the increasing trend about dialogue modeling, response selection and generation have been widely studied in the NLP community.
In order to further evaluate the current capability of the machine learning models, a benchmark dataset was proposed in the seventh Dialog System Technology Challenge (DSTC7) \cite{DSTC7}, where the task is to select the most probable response given a partial conversation.
To simulate real world scenarios, several variants of selections are investigated in this task: 1) selecting from 100 candidates, 2) selecting from 120,000 candidates, 3) selecting multiple answers, and 4) there may be no answer.
Some subtasks are much more difficult than the original setting.
In addition, the ability of generalization should be examined; hence, two datasets, Ubuntu IRC dialogs \cite{kummerfeld2018analyzing} and course advising corpus, are utilized for the experiments.
These datasets have very different properties: (1) the Ubuntu dataset includes dialogues from Ubuntu IRC channel that aim to solve technical problems, and (2) the advising dataset is constituted with conversations between a student and an advisor, where the advisor helps the student about course taking.
Compared with the advising dataset, utterances in a dialogue from Ubuntu data are more coherent, so selecting the next sentence may require understanding of previous dialogue turns.
On the contrary, the topic in a dialogue from advising data may change frequently, but the information access behavior is more goal-oriented.
How much information in the dialogue contexts should be considered for sentence selection in these two datasets may be different, so how to effectively utilize the information is challenging and salient.
In sum, the challenge covers a wide range of scenarios in real-world applications and serves as a set of benchmark experiments for evaluating dialogue response selection models.

The nature of a conversation is different from general articles, because opinion, topic and meaning of some terms may change as the dialogue proceeds.
Meanwhile, a conversation comprises utterances spoken by the participants, where each utterance is short and having clear boundary. 
Especially, one utterance is often a response of its previous utterance; 
therefore we could expect high dependency between every two consecutive utterances in a conversation.

Modeling dependency over utterances should be helpful for understanding conversations; however, the methods of modeling such dependency have not been widely explored. 
In this work, we propose \emph{highway recurrent transformer} to explicitly model not only intra-utterance but also inter-utterance dependency over the dialogue structure.
The intra-utterance dependency is modeled with Transformer encoder block proposed in \cite{vaswani2017attention}
, while the inter-utterance dependency is modeled by using the proposed highway attention recurrently.
Specifically, the highway attention is a modified version of multi-head attention, designed to have the ability to utilize the information from contexts while preserving the meaning of the current utterance.
Experiments show that highway recurrent transformer model is effective on the response selection task; furthermore, the proposed model can also generalize to other retrieval-based tasks.

\begin{figure*}[t!]
  \centering
  \includegraphics[width=0.8\linewidth]{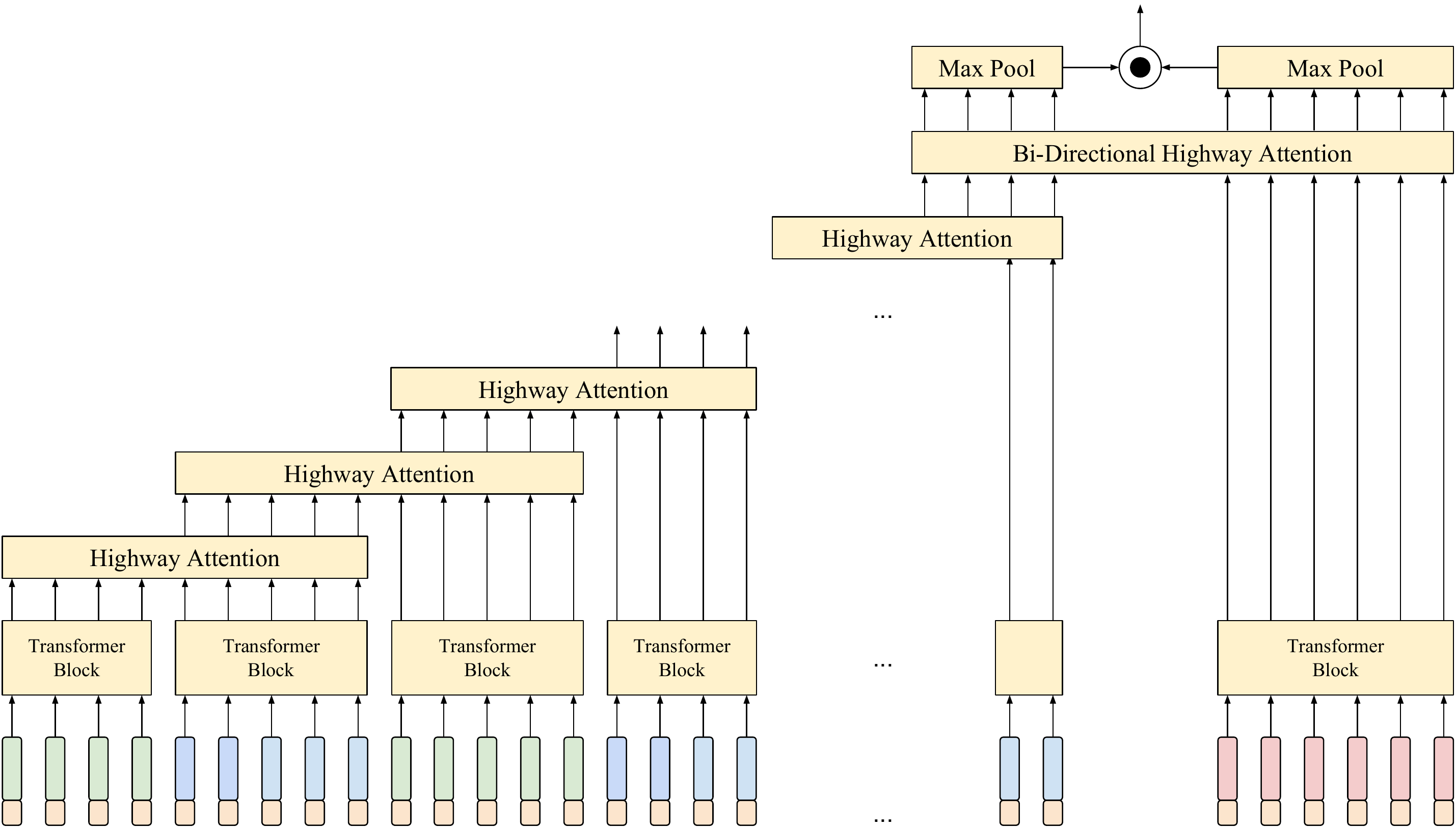}
  \caption{The illustration of the proposed highway recurrent transformer. In the bottom left part, the blue and green rounded rectangles represent word vectors in the partial conversation $U$ spoken by two different speakers. 
  The pink rounded rectangles at the bottom right corner represent word vectors in a candidate $x$. Both two sequences of vectors are augmented with extra features (light orange part beneath the vectors).}
  \label{fig:recurrent-transformer}
\end{figure*}

\section{Related Work}
Previously, another large-scale dataset, Ubuntu Dialog Corpus \cite{lowe2015ubuntu} was proposed, and a large number of approaches have been applied to this data.
One category of the approaches encodes the given partial conversation and candidate sentences into vectors separately, and then selects the answer by matching the vectors.
In this category, LSTM and CNN were applied to encode the dialogues and response candidates \cite{lowe2015ubuntu,kadlec2015improved,hochreiter1997long,lecun1998gradient}. 
Moreover, \citeauthor{zhou2016multi} used GRU and CNN to utilize a hierarchical structure for obtaining word-level and conversation-level representations.
Another category of the approaches focuses on explicitly matching the conversation and the candidates instead of encoding them into vectors.
In this category, \citeauthor{wu2017sequential} matched the GRU-encoded words in the utterances and candidates with an attention mechanism \cite{bahdanau2014neural}.
\citeauthor{zhang2018modeling} modeled not only the relation between utterances and responses but also the relation between utterances and the last utterance in the given partial conversation. \citeauthor{zhou2018multi} formed a 3D similarity matrix by stacking the matching matrices between words in the utterances and each candidate, and then 3D convolution is used to calculate the score for each candidate.
Among the two categories, only few works considered the relation between utteraces~\cite{zhang2018modeling,zhou2018multi}, but none applied the attention for this task.
Specifically, \citeauthor{zhang2018modeling} only modeled the utterance relation to the last one, while \citeauthor{zhou2018multi} modeled the relation between utterances implicitly with a convolution operation.

\section{Task Description}

In the response selection challenge, a partial conversation and a set of response candidates are provided, and the model needs to select one response from the candidates set.
The partial conversation consists of $l$ utterances: $U: \{u_1, u_2, \cdots, u_l\}$, an utterance is a sequence of words, $i$-th utterance is denoted as  $u_i: \{w_{i,0}^U, w_{i,1}^U, w_{i,2}^U, \cdots, w_{i,n_i}^U \}$.
Each speaker participating in the conversation is given a special token, say \texttt{<speaker1>}, \texttt{<speaker2>}, and the special token is prepended to the utterances from that speaker.
A candidate set consisting of $k$ candidates is denoted as $X: \{x_1, x_2, \cdots, x_k \}$, each candidate is a sequence of words $x_j: \{ w_{j,1}^X, w_{j,2}^X, \cdots, w_{j,m_j}^X \}$.
For some datasets, some knowledge-grounded features of a word $w$ are also available, denoted as $F(w)$.
Among the candidates, none or some would be the correct responses, the labels indicating if the candidates are  correct answers are denoted as $Y: \{y_1, y_2, \cdots, y_k\}$.

\section{Highway Recurrent Transformer}
We propose Highway Recurrent Transformer to model both the intra-utterance and inter-utterance dependency, which is composed of two main component: Transformer \cite{vaswani2017attention} and highway attention. The whole model architecture is illustrated in Figure \ref{fig:recurrent-transformer}.

\subsection{Word Feature Augmentation}
\label{ssec:word_feature}
Words in the utterances of a conversation and candidates are first converted into their word embeddings, and the embeddings are augmented with some extra knowledge-grounded features if such features are available.
We denote the sequences of words with extra features in the context, candidates as
\begin{align}
    & \tilde{U}: \{\tilde{u}_1, \tilde{u}_2, \cdots, \tilde{u}_l \}, \nonumber \\
    & \tilde{X}: \{\tilde{x}_1, \tilde{x}_2, \cdots, \tilde{x}_k \}, \nonumber
\end{align}
each word embeddings in the sequences is concatenated with features
\begin{align}
    \tilde{w} = [w; F(w)], \nonumber
\end{align}
where $F(w)$ is a vector of knowledge grounded features depending on the dataset.

\subsection{Transformer Encoder Block}
\begin{figure}[t!]
  \centering
  \includegraphics[width=0.5\linewidth]{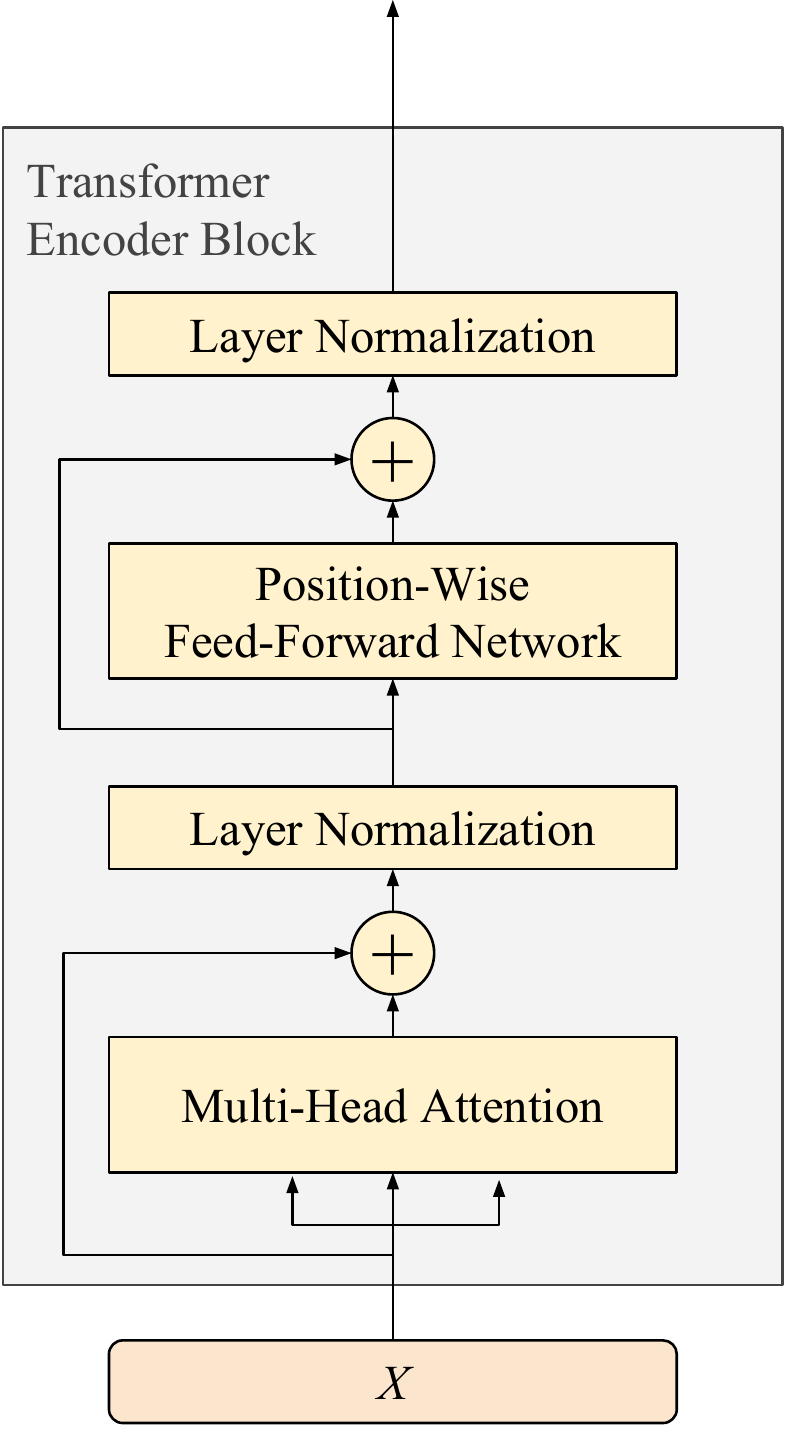}
  \vspace{-2mm}
  \caption{The illustration of the encoding part in the highway recurrent transformer. The input of multi-head attention module includes key, query, and value sequences; the bottom branches implies that $X$ is fed as the three parameters at the same time.}
  \label{fig:transformer-encoder-block}
  \vspace{-1mm}
\end{figure}

A transformer encoder block (Figure \ref{fig:transformer-encoder-block}) proposed in \cite{vaswani2017attention} consists of a multi-head attention layer and a position-wise feed-forward network, residual connection and layer normalization are used to connect the two components, details are specified as follows.

\begin{figure}[t!]
  \centering
  \includegraphics[width=1\linewidth]{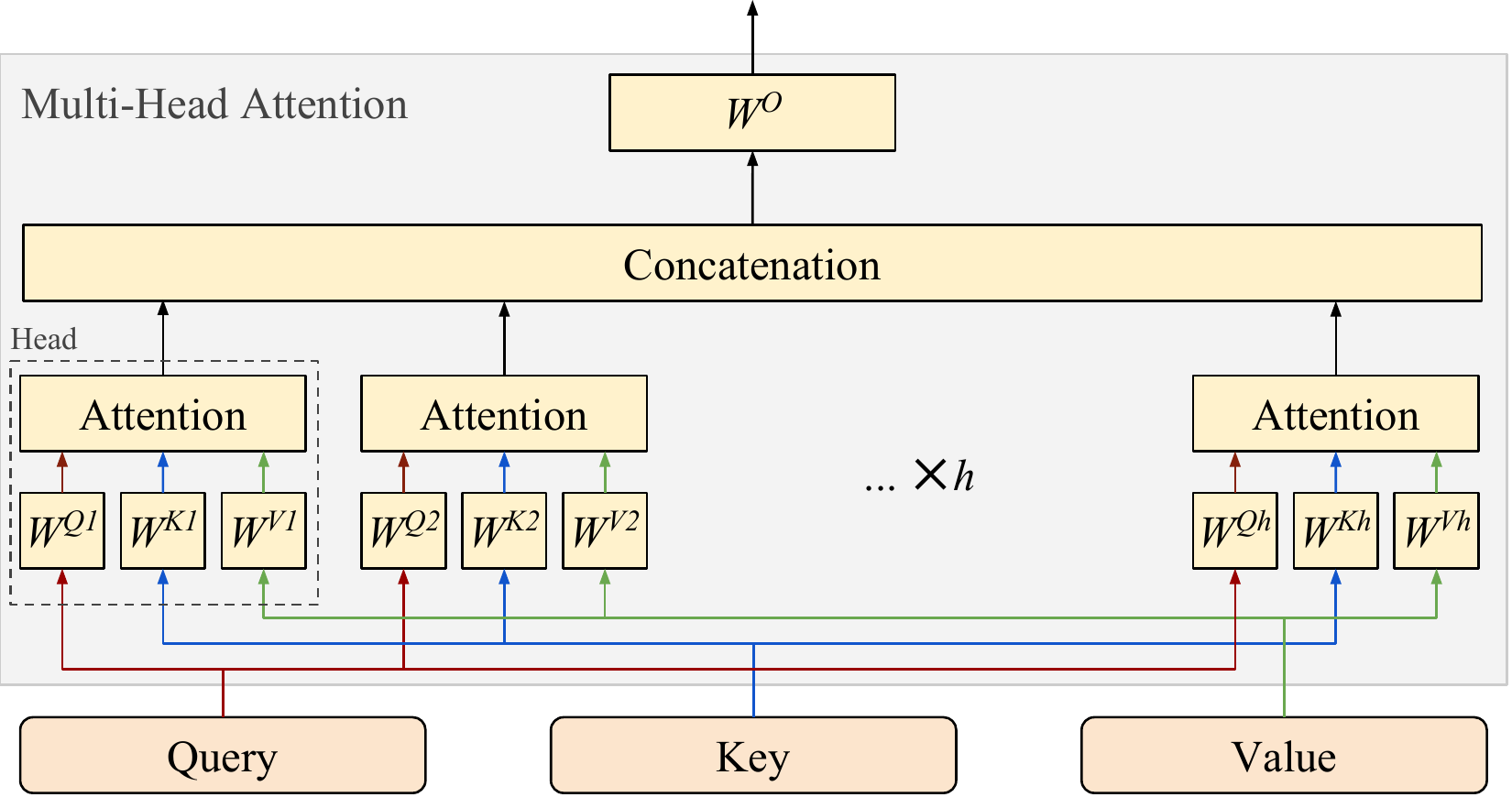}
  \vspace{-2mm}
  \caption{The illustration of the multi-head attention layer. Note that the input \textit{Query}, \textit{Key}, \textit{Value} and the output are sequences of vectors.}
  \label{fig:multi-head-attention}
  \vspace{-1mm}
\end{figure}
\subsubsection{Multi-Head Attention Layer}

A multi-head attention \cite{vaswani2017attention} (Figure \ref{fig:multi-head-attention}) consists of the heads of attention, each head performs linear transformation before performing attention operation; with different sets of trainable parameters, each attention head potentially models different relationship between two sequences.
Specifically, the inputs of the multi-head attention layer are three sequences of vectors: \emph{query} $Q \in \mathbb{R}^{l_1 \times d_f}$, \emph{key} $K  \in \mathbb{R}^{l_2 \times d_f}$, \emph{value} $V \in \mathbb{R}^{l_2 \times d_f}$, where $l_1, l_2$ are the length of the first and second sequence respectively. 
Then for the $h$-th head, three weight matrices $W^{Qh}, W^{Kh}, W^{Vh} \in \mathbb{R}^{d_f \times d_p}$ are used to project the three inputs to a lower dimension $d_p$, and then an attention function is performed
\begin{equation}
\label{eq:multi-head-attn-trans}
    A^h = \mathrm{Attention}(Q^h, K^h , V^h), \nonumber
\end{equation}
where $Q^h = QW^{Qh}, K^h = KW^{Kh}, V^h = VW^{Vh}$.
The attention function generates a vector for each vector in the query sequence $Q$. Let the outputs of the attention function be $A^h \in \mathbb{R}^{l_1 \times d_p}$, which is weighted sum of value $V$ based on similarity matrices $S^h$.
For $a = 1, 2, \cdots, l_1$, the $a$-th output is calculated as below:
\begin{align}
  &S^h = Q^h (K^h)^T, \nonumber \\
  &A^h_a = \sum_{p=1}^{l_2} \frac{\exp s^h_{a, p} }{\sum_{t=1}^{l_2} \exp s^h_{a, t}} V^h_{p}, \label{eq:multi-head-attn-a} 
\end{align}
where $s$ are the similarity scores in the similarity matrix $S^h$.
Then the output of the multi-head attention is the linear transformed concatenation of the outputs from attention heads:
\begin{equation}
  \mathrm{MultiHead}(Q, K, V) = \mathrm{Concat}(A^1, A^2, \cdots, A^H) W^O \nonumber
  \label{eq:multi-head-attn}
\end{equation}
where $H$ is the number of heads, and $W^O \in W^{H \cdot d_p \times d_f}$ is a trained weight matrix.

\subsubsection{Position-Wise Feed-Forward Network}

The position-wise feed-forward network (FFN) transforms each vector in a sequence identically as follows:
\begin{align}
  \mathrm{FFN}(X) = \max(0, XW_1 + b_1)W_2 + b_2 \nonumber
\end{align}

\subsubsection{Residual Connection and Layer Normalization}
Then the above two components are connected with residual connection and layer normalization \cite{ba2016layer}:
\begin{equation}
\label{eq:transformer-block}
\begin{aligned}
  & \mathrm{ResiNorm}(f, X) = \mathrm{LayerNorm}(X + f(X)), \\
  & \mathrm{TransformerBlock}(X) = \nonumber \\
  & \mathrm{ResiNorm}(\mathrm{FFN}, \mathrm{ResiNorm}(\mathrm{MultiHead}, (X, X, X))).
\end{aligned}
\end{equation}
Note that here we use the same sequence for the query, key, value arguments of the mult-head attention for self-attention.

\subsection{Highway Attention}

\begin{figure}[t!]
  \centering
  \includegraphics[width=1\linewidth]{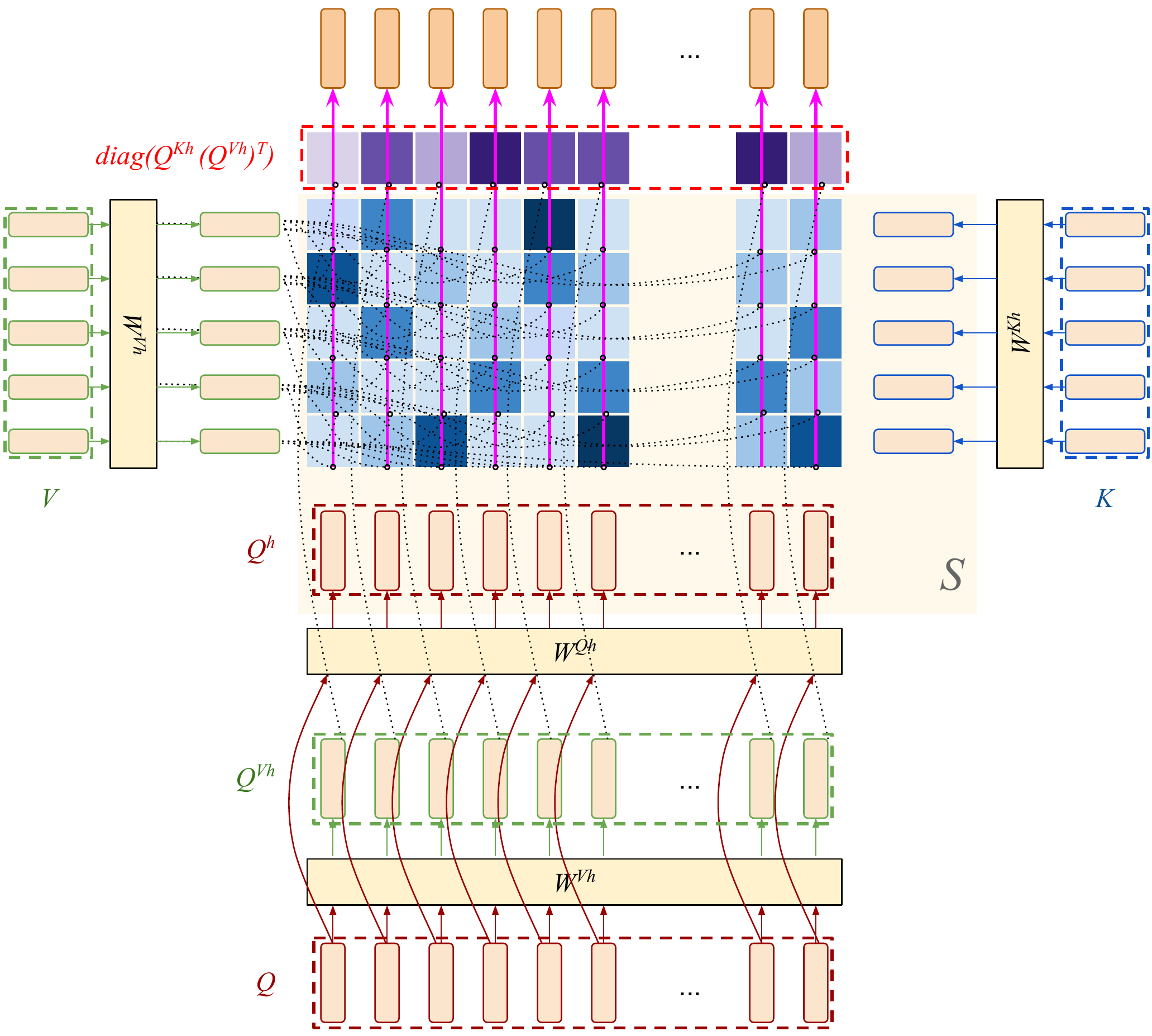}
  \caption{Highway Attention. The attention values $S$ (blue blocks) are the inner product of $Q^h$ and $K^h$ (rounded rectangles with red border and blue border respectively). The dotted curve from one vector (rounded rectangle) to the attention value (blue block) denotes that the vector is weighted by the value. So the output vectors (top round rectangles with red borders) are the summation of the vectors $V, Q^{Vh}$ (rounded rectangles with green border) weighted by the attention values. }
  \label{fig:highway-attention}
\end{figure}

Motivated by Highway Networks \cite{srivastava2015highway}, we propose a modified version of multi-head attention, the highway attention (Figure \ref{fig:highway-attention}), in which attention also acts as a highway preserving information from the lower layer.
The highway is achieved by performing attention on the query vector itself in addition to the key/value sequence. 
Specifically, given query $Q \in \mathbb{R}^{l_1 \times d_f}$, key $K  \in \mathbb{R}^{l_2 \times d_f}$, value $V \in \mathbb{R}^{l_2 \times d_f}$, in addition to linear transformation defined in equation (\ref{eq:multi-head-attn-trans}), we also transform the query sequence $Q$ with $W^{Kh}$ and $W^{Vh}$ into additional features:
\begin{align}
  \begin{cases}
      Q^h = QW^{Qh}, \\
      K^h = (K + b_{co}) W^{Kh}, \\
      V^h = VW^{Vh}, \\
      Q^{Kh} = (Q + b_{self}) W^{Kh}, \\
      Q^{Vh} = Q W^{Vh}. \\
  \end{cases}
\end{align}
The bias vectors $b_{co}$ and $b_{self}$ are added to each vector in sequence $K$ and $Q$ respectively. They are added for two reasons:
First, they encode the information of the key's source, whether the key is from the key sequence $K$ or from the query sequence $Q$, it may be crucial since the calculation of the similarity matrix $S$ is position-independent.
Secondly, the tendency of co-attention or self-attention could be modeled; in other words, it may learn some prior knowledge which is independent of current data samples.
Then the similarity matrices $S \in \mathbb{R}^{l_1 \times l_2}$ and $S_{self} \in \mathbb{R}^{l_1}$ are calculated:
\begin{align}
  S^h &= Q^h (K^h)^T , \nonumber \\
  S^h_{self} &= \mathrm{diag}(Q^{h} (Q^{Kh})^T). \nonumber
\end{align}
Note that here we only take the diagonal of the self-attention similarity matrix $S^h_{self}$ to measure the tendency toward focus on itself for each vector.
Hence (\ref{eq:multi-head-attn-a}) could be expanded as
\begin{align}
  \label{eq:highway-attn-self}
  A^h_a =& \sum_{p=1}^{l_2} \frac{\exp s^h_{a, p} }{\sum_{t=1}^{l_2} \exp s^h_{a, t} + \exp s^h_{self, a}} V^h_{p} \\ 
         & + \frac{s^h_{self, a} }{\sum_{t=1}^{l_2} \exp s^h_{a, t} + \exp s^h_{self,a}} Q^{Vh}_a. \nonumber
\end{align}
Namely, $A^h_a$ is the weighted sum of $Q^{Vh}_a$ and vectors in sequence $V^h$.
When $S_{self}^h$ is larger, $A^h$ will retain more information of $Q^h$. If we treat the attention mechanism as a transformation that transforms vector sequence $Q$ into vector sequence $A$, then the design here provides a highway to preserve lower layer information.
Finally, as in equation (\ref{eq:multi-head-attn}), we define the Highway Attention as linear transformation of the concatenation of the heads:
\begin{equation}
    \mathrm{HighwayAttn}(Q, K, V) = \mathrm{Concat}(A^1, \cdots, A^H) W^O.\nonumber
\end{equation}

\subsubsection{Recurrence}
Each utterance $\tilde{u}_i$ is encoded with a transformer block that shares the parameter of the ones in (\ref{eq:transformer-block}) with the positional encoding:
\begin{align}
  v_i^U = \mathrm{TransformerBlock}(\tilde{u}_i + \mathrm{PE}), \nonumber 
\end{align}
where $\mathrm{PE}$ represents positional encoding.

Because the current utterances may refer to the instances mentioned in the previous utterance, the highway attention is applied recurrently to route the information from the previous utterances $a_{i-1}^U$ to the current one $v_i^U$ and further infer the attended output $a_i^U$:
\begin{align}
    a_i^U =
    \begin{cases}
      v_1^U & \text{if $i = 1$}, \\
      \mathrm{HighwayAttn}(v_i^U, a_{i-1}^U)  & \text{if $i > 1$}. \nonumber \\
    \end{cases}
\end{align}

From another perspective, if we view the outputs of the highway attention for a utterance as the ``memory'' that represents the dialogue state till now, the Highway Attention is hence very similar to the update gate of GRU \cite{cho2014learning}. Equation (\ref{eq:highway-attn-self}) decides how much information is read into the memory, similar to the update gate of GRU. 

\subsection{Candidate Selection}
Similarly, each candidate $\tilde{x}_j$ is encoded with a transformer block with the positional encoding:
\begin{align}
  v_j^X = \mathrm{TransformerBlock}(\tilde{x}_j + \mathrm{PE}). \nonumber
\end{align}
The bi-directional highway attention is applied to model the relation between conversations and candidates in two directions.
\begin{eqnarray}
 \label{eq:bi-attention-last-1}
  \tilde{v}_j^X &=& \mathrm{HighwayAttn}(v_j^X, a_l^U)  \\
  \label{eq:bi-attention-last-2}
  \tilde{a}^U &=& \mathrm{HighwayAttn}(a_l^U, v_j^X) 
\end{eqnarray}
where $a_l^U$ represents the most recent output of highway attention from utterances $U$.
Note that the parameters of the two highway attention blocks are shared. 

In another similar model, \textit{highway recurrent transformer-all},
all utterances are considered when applying bidirectional highway attention, so (\ref{eq:bi-attention-last-1}) and (\ref{eq:bi-attention-last-2}) are replaced with
\begin{equation}
\begin{aligned}
\label{eq:bi-attention-all}
  \tilde{v}_j^X &= \mathrm{HighwayAttn}(v_j^X, [a_1^U; a_2^U; \cdots; a_l^U]),  \nonumber \\
  \tilde{a}^U &= \mathrm{HighwayAttn}([a_1^U; a_2^U; \cdots; a_l^U], v_j^X). 
\end{aligned}
\end{equation}

For the submitted system, the attention mechanism is used to condense two sequences into two vectors by weighted sum over feature sequences:
\begin{align} 
     \alpha_k &= w^T \tilde{v}^X_{j,k}, \quad r_j^X = \sum_k \alpha_k \tilde{v}^X_{j,k},       \nonumber    \\
    \alpha_k &= w^T \tilde{a}^U_{k},  \quad r^U = \sum_k \alpha_k \tilde{a}^U_{k}, \nonumber
\end{align}
where $w$ is a trainable weight vector.
The score of a candidate $x_j$ is calculated as $s_j = r_j^X \cdot r^U$.
However, we afterwards find max pooling over dimensions is more effective:
\begin{equation}
  r_j^X = \max(\tilde{v}_j^X), r^U = \max(\tilde{a}^U),  \nonumber
\end{equation}
therefore the max-pooling method is conducted in the subsequent experiments in this paper.

We trained the submitted system with the \textit{ranking loss}, which gives the additional penalty if the lowest score among the positive samples' scores is not greater than the highest one among the negative ones' by a margin:
\begin{equation}
\begin{aligned}
 \mathrm{LSE}(S) =& \log \sum_{s \in S} \exp\{ s \}, \\
  L_{rank}(U, X, Y) = &\max(0, \mathrm{LSE}(\{s_j | y_j = 0 \} \nonumber \\
  + &\mathrm{LSE}(\{- s_j | y_j = 1 \}) + \gamma),\\
\end{aligned}
\end{equation}
where $S$ is the set of scores of candidates and $\mathrm{LSE}(.)$ is a smooth approximation of maximum function.
Afterwards, we find it does not outperform binary cross entropy, so in the experiments in this paper, we utilize the binary cross entropy function as the objective:
\begin{align}
  L(U, X, Y) = \sum_{j=1}^k y_j \log \sigma(s_j) + (1 - y_j) \log (1 - \sigma(s_j)) \nonumber
\end{align}

\section{Experiments}

\subsection{Dataset}

DSTC7-Track1 contains two goal-oriented dialogue datasets --- 1) Ubuntu data and 2) Advising data.
There are five subtasks in this challenge, where this paper focuses on the subtask 1, 3 and 4, because the same model architecture can be applied to these subtasks.
Here we briefly describe the settings for each subtask:
\begin{compactitem}
  \item Subtask 1: There are 100 response candidates for each dialogue, and only one is correct.
  \item Subtask 3: There are 100 response candidates for each dialogue. The number of correct responses is between 1 and 5. Multiple correct responses are generated using paraphrases.
  \item Subtask 4: There are 100 response candidates for each dialogue, and an additional candidate \textit{no answer} should also be considered. The number of correct responses is either 0 or 1.
\end{compactitem}

\subsection{Settings}

We train and evaluate our model on the Ubuntu and Advising datasets provided by DSTC7 \cite{DSTC7} track 1. Both the datasets are tokenized with Spacy \cite{spacy2} and pre-trained word embeddings FastText \cite{bojanowski2017enriching} are used. For the Advising dataset, in the preprocessing phase, course numbers are normalized to a uniform format (e.g. CS1234n). And we define the domain specific binary features with the provided suggested courses list and the prior taken courses list:
\begin{align}
    F(w) = [F^{prior}(w),F^{suggested}(w)]
\end{align}
where $F^{prior}(w)$ and $F^{suggested}(w)$ are $1$ if and only if $w$ is a course number and the course is in the prior taken courses list or the suggested courses list respectively. For the position-wise feed forward function in the highway recurrent transformer model, 512 is used as the hidden layer dimension, and each utterance is encoded by 2 Transformer encoder blocks. The model is trained by sampling negative candidates so the total number of candidates is 10 for each sample. The whole model is optimized with Adam optimizer \cite{kingma2014adam} with learning rate 0.0001.

\begin{table*}[t!]
  \centering
  \small
\begin{tabular}{ll|ccc|ccc}
  \toprule
  && \multicolumn{3}{c|}{Ubuntu} & \multicolumn{3}{c}{Advising} \\
  && Recall@10  & MRR & Average & Recall@10  & MRR  & Average   \\
  \midrule
  \multirow{7}{*}{Subtask1}
  &Dual LSTM               & 62.6 / 58.7 & 36.23 / 35.37 & 49.39 / 47.03 & 62.8 / 39.8 & 31.36 / 15.71 & 47.14 / 27.76 \\
  &Hierarchical LSTM       & 65.3 / 57.5 & 37.83 / 34.54 & 51.56 / 46.02 & 64.4 / 47.8 & 32.32 / 23.84 & 48.42 / 35.82 \\
  &Transformer             & 64.8 / 60.6 & 40.55 / 36.06 & 52.68 / 48.33 & 71.6 / 52.4 & 40.19 / 25.30 & 55.90 / 38.85 \\
  &Transformer-Last        & 62.8 / 49.1 & 35.40 / 28.08 & 49.10 / 38.59 & 70.4 / 52.6 & 39.23 / 25.12 & 54.81 / 38.86 \\
  \cmidrule{2-8}
  &Highway Recur. Trans.-A & 74.24 / 66.2 & 46.01 / 40.40 & 60.12 / 53.30 & 71.6 / 48.4 & 39.30 / 21.43 & 55.45 / 34.91 \\  
  &Highway Recur. Trans.-L & 74.06 / 67.0 & 45.18 / 40.19 & 59.62 / 53.60 & 69.8 / 51.6 & 42.32 / 25.36 & 56.07 / 38.48 \\
  &Ensemble                & \bf 89.62 / 67.9 & \bf 66.37 / 43.52 & \bf 77.99 / 55.71 & \bf 77.0 / 54.8 & \bf 39.98 / 27.89 & \bf 58.49 / 41.34 \\
  \midrule
  \multirow{7}{*}{Subtask3}
  &Dual LSTM             & - & - & - & 65.0 / 47.4 & 43.43 / 22.41 & 54.25 / 34.91 \\
  &Hierarchical LSTM     & - & - & - & 73.4 / 52.2 & 52.41 / 25.73 & 62.91 / 38.96 \\
  &Transformer           & - & - & - & 79.2 / 60.6 & 55.00 / 29.89 & 67.10 / 45.25 \\
  &Transformer-Last      & - & - & - & \bf 81.0 / 61.2 & 56.42 / 31.69 & 68.71 / {\bf 46.45} \\
  \cmidrule{2-8}
  &Highway Recur. Trans.-A & - & - & - & 80.8 / 57.2 & 53.54 / 27.36 & 67.17 / 42.28 \\
  &Highway Recur. Trans.-L & - & - & - & 70.8 / 54.6 & 45.52 / 29.08 & 58.16 / 41.84 \\
  &Ensemble \& Fine Tune      & - & - & - & 81.0 / 60.4 & \bf 57.15 / 31.71 & {\bf 69.08} / 46.06 \\
  \midrule
  \multirow{7}{*}{Subtask4}
  &Dual LSTM             & 61.9 / 69.1 & 34.72 / 38.94 & 48.31 / 54.02 & 64.0 / 43.8 & 28.19 / 18.20 & 46.09 / 31.00 \\
  &Hierarchical LSTM     & 55.5 / 57.9 & 32.73 / 34.44 & 44.14 / 46.17 & 57.0 / 40.6 & 31.25 / 17.28 & 44.08 / 28.94 \\
  &Transformer           & 70.4 / 75.3 & 40.77 / 46.29 & 55.60 / 60.79 & 69.6 / 52.2 & 32.62 / 22.16 & 51.11 / 37.18 \\
  &Transformer-Last      & 60.6 / 62.4 & 34.35 / 36.84 & 47.47 / 49.62 & \bf 68.7 / 56.8 & {\bf 33.92} / 24.42 & \bf 51.31 / 40.61 \\
  \cmidrule{2-8}
  &Highway Recur. Trans.-A & 71.1 / {\bf 75.7} & \bf 41.37 / 46.51 & \bf 56.23 / 61.11 & 60.6 / 40.4 & 29.30 / 16.28 & 44.95 / 28.34 \\
  &Highway Recur. Trans.-L & 68.8 / 73.3 & 38.56 / 41.72 & 53.70 / 57.51 & 57.8 / 39.2 & 23.54 / 25.18 & 40.67 / 32.19 \\
  &Ensemble                & {\bf 71.2} / 72.9 & 40.89 / 43.37 & 56.04 / 58.13 & 64.2 / 46.4 & 33.01 / {\bf 27.09} & 48.61 / 36.75 \\
  \bottomrule
\end{tabular}
\vspace{-1.5mm}
\caption{Performance of the different models on the validation / test sets.}
\vspace{-1mm}
\label{table:performance}
\end{table*}

\subsection{Baseline Models}

We compare our model with following baseline models:
\begin{compactitem}
    \item Dual LSTM~\cite{lowe2015ubuntu}: uses two LSTMs to encode the conversation and the candidates into two vectors, and the inner product is used to select the candidate.
    \item Hierarchical LSTM: is based on the encoder in HRED \cite{serban2016building} for encoding the conversations, where one LSTM is used to encode an utterance or a candidate into one vector as its utterance representation, and then the second LSTM encodes the utterance-level representations into a conversation-level representation. Finally, the candidate is selected based on the inner product of its representation and the conversation-level representation.
    \item Transformer: The utterances are concatenated as a single sequence and then both the sequence and the candidate encoded by layers of transformer encoder blocks. Then bi-directional highway attention and max pooling are applied and two vectors that represent the conversation and the candidate are obtained. Their inner product is used to for selection.
    \item Transformer-Last: Same as the transformer described above, but only the last utterance is fed into the model.
\end{compactitem}

\subsection{Ubuntu Results}

The performance comparison is shown in Table \ref{table:performance}.
It is obvious that on the Ubuntu data, the proposed highway recurrent transformer outperforms the transformer baseline, and the highway recurrent transformer-last also outperforms the transformer-last.
Especially for subtask 1, both highway recurrent transformer and highway recurrent transformer-last significantly outperform the transformer without recurrent highway attention.
For subtask 4, the highway recurrent transformer-last also obtains comparable performance with the transformer, but still worse than the highway recurrent transformer-all.
Note that the highway recurrent transformer models the relation between conversations and the candidates in the lower layer, and the relation is modeled by the bi-directional highway attention layer at almost the last layer.
In other words, our model encodes the partial conversation in vectors independent of the candidates, and the computation cost for scoring one candidate is only the bi-directional highway attention and the inner product.
It can be a great advantage over other approaches that model the relation between conversations and candidates in the lower layer.
The highway recurrent transformer-last further reduces the computation cost required to score candidates by considering only the output of the last utterance.
Therefore, the performance loss incurred by the highway recurrent transformer-last can be seen as the trade-off between the accuracy and the efficiency.

\subsection{Advising Results}
In the advising data, the advantage of the recurrent structure is not as significant as in the Ubuntu data.
From Table \ref{table:performance}, we can also see that for the subtask 1 and 3 on advising data, transformer-all and transformer-last obtain similar performance, implying that utterances prior to the last one have little useful information to predict the next one. 
That may indicate why the recurrent highway attention is not as useful on the advising dataset. 
Nevertheless, it is surprising that the ensemble of highway recurrent transformer-last and the hierarchical LSTM leads to significant performance boost compared with either one of two single models.
The results demonstrate that the proposed highway recurrent transformer-last model may have some complementary advantages the hierarchical LSTM does not have. 
In sum, the proposed model achieves the improvement for the subtask 1 and comparable performance for the subtasks 3 and 4 with the transformer model for advising dialogues.

\subsection{Official Evaluation}

In the official evaluation, the submitted systems are not the optimal ones. 
All submitted systems use 1 layer of the transformer encoder block to encode the utterances.
For subtasks 1 and 4, we ensemble the results of the hierarchical LSTM and our highway recurrent transformer-last by summing the scores of all candidates before applying the sigmoid function.
As for subtask 3, in addition to the summation trick above, we further fine-tune the whole model.
To sum up, the submitted systems include:
\begin{compactitem}
    \item Subtask 1: Ensemble
    \item Subtask 3: HierRNN, Highway Recurrent Transformer-Last, Ensemble \& Fine-Tuning
    \item Subtask 4: HierRNN, Ensemble
\end{compactitem}
Our submitted systems achieve either the 2nd place or the 3rd place for the above three subtasks for both Ubuntu and advising data by combining with another attention-based model, demonstrating the effectiveness of our model.

\section{Conclusion}
This paper proposes a highway recurrent transformer that effectively models information from multiple levels, including utterance-level and conversation-level, via a highway attention mechanism.
The experiments of DSTC7 empirically demonstrate the superior capability of estimating the relation between the dialogue and the response and further selecting the proper response given the dialogue contexts.
Compared with the state-of-the-art transformer models, our proposed model achieves improved performance.
The proposed highway recurrent transformer can be investigated for other tasks in the future.


\bibliography{main}

\begin{thebibliography}{}

\bibitem[\protect\citeauthoryear{Ba, Kiros, and Hinton}{2016}]{ba2016layer}
Ba, J.~L.; Kiros, J.~R.; and Hinton, G.~E.
\newblock 2016.
\newblock Layer normalization.
\newblock {\em arXiv preprint arXiv:1607.06450}.

\bibitem[\protect\citeauthoryear{Bahdanau, Cho, and
  Bengio}{2014}]{bahdanau2014neural}
Bahdanau, D.; Cho, K.; and Bengio, Y.
\newblock 2014.
\newblock Neural machine translation by jointly learning to align and
  translate.
\newblock {\em arXiv preprint arXiv:1409.0473}.

\bibitem[\protect\citeauthoryear{Bojanowski \bgroup et al\mbox.\egroup
  }{2017}]{bojanowski2017enriching}
Bojanowski, P.; Grave, E.; Joulin, A.; and Mikolov, T.
\newblock 2017.
\newblock Enriching word vectors with subword information.
\newblock {\em Transactions of the Association for Computational Linguistics}
  5:135--146.

\bibitem[\protect\citeauthoryear{Cho \bgroup et al\mbox.\egroup
  }{2014}]{cho2014learning}
Cho, K.; Van~Merri{\"e}nboer, B.; Gulcehre, C.; Bahdanau, D.; Bougares, F.;
  Schwenk, H.; and Bengio, Y.
\newblock 2014.
\newblock Learning phrase representations using rnn encoder-decoder for
  statistical machine translation.
\newblock {\em arXiv preprint arXiv:1406.1078}.

\bibitem[\protect\citeauthoryear{Hochreiter and
  Schmidhuber}{1997}]{hochreiter1997long}
Hochreiter, S., and Schmidhuber, J.
\newblock 1997.
\newblock Long short-term memory.
\newblock {\em Neural computation} 9(8):1735--1780.

\bibitem[\protect\citeauthoryear{Honnibal and Montani}{2017}]{spacy2}
Honnibal, M., and Montani, I.
\newblock 2017.
\newblock spacy 2: Natural language understanding with bloom embeddings,
  convolutional neural networks and incremental parsing.
\newblock {\em To appear}.

\bibitem[\protect\citeauthoryear{Kadlec, Schmid, and
  Kleindienst}{2015}]{kadlec2015improved}
Kadlec, R.; Schmid, M.; and Kleindienst, J.
\newblock 2015.
\newblock Improved deep learning baselines for ubuntu corpus dialogs.
\newblock {\em arXiv preprint arXiv:1510.03753}.

\bibitem[\protect\citeauthoryear{Kingma and Ba}{2014}]{kingma2014adam}
Kingma, D.~P., and Ba, J.
\newblock 2014.
\newblock Adam: A method for stochastic optimization.
\newblock {\em arXiv preprint arXiv:1412.6980}.

\bibitem[\protect\citeauthoryear{Kummerfeld \bgroup et al\mbox.\egroup
  }{2018}]{kummerfeld2018analyzing}
Kummerfeld, J.~K.; Gouravajhala, S.~R.; Peper, J.; Athreya, V.; Gunasekara, C.;
  Ganhotra, J.; Patel, S.~S.; Polymenakos, L.; and Lasecki, W.~S.
\newblock 2018.
\newblock Analyzing assumptions in conversation disentanglement research
  through the lens of a new dataset and model.
\newblock {\em arXiv preprint arXiv:1810.11118}.

\bibitem[\protect\citeauthoryear{LeCun \bgroup et al\mbox.\egroup
  }{1998}]{lecun1998gradient}
LeCun, Y.; Bottou, L.; Bengio, Y.; and Haffner, P.
\newblock 1998.
\newblock Gradient-based learning applied to document recognition.
\newblock {\em Proceedings of the IEEE} 86(11):2278--2324.

\bibitem[\protect\citeauthoryear{Lowe \bgroup et al\mbox.\egroup
  }{2015}]{lowe2015ubuntu}
Lowe, R.; Pow, N.; Serban, I.; and Pineau, J.
\newblock 2015.
\newblock The ubuntu dialogue corpus: A large dataset for research in
  unstructured multi-turn dialogue systems.
\newblock {\em arXiv preprint arXiv:1506.08909}.

\bibitem[\protect\citeauthoryear{Serban \bgroup et al\mbox.\egroup
  }{2016}]{serban2016building}
Serban, I.~V.; Sordoni, A.; Bengio, Y.; Courville, A.~C.; and Pineau, J.
\newblock 2016.
\newblock Building end-to-end dialogue systems using generative hierarchical
  neural network models.
\newblock In {\em AAAI}, volume~16,  3776--3784.

\bibitem[\protect\citeauthoryear{Srivastava, Greff, and
  Schmidhuber}{2015}]{srivastava2015highway}
Srivastava, R.~K.; Greff, K.; and Schmidhuber, J.
\newblock 2015.
\newblock Highway networks.
\newblock {\em arXiv preprint arXiv:1505.00387}.

\bibitem[\protect\citeauthoryear{Vaswani \bgroup et al\mbox.\egroup
  }{2017}]{vaswani2017attention}
Vaswani, A.; Shazeer, N.; Parmar, N.; Uszkoreit, J.; Jones, L.; Gomez, A.~N.;
  Kaiser, {\L}.; and Polosukhin, I.
\newblock 2017.
\newblock Attention is all you need.
\newblock In {\em Advances in Neural Information Processing Systems},
  5998--6008.

\bibitem[\protect\citeauthoryear{Wu \bgroup et al\mbox.\egroup
  }{2017}]{wu2017sequential}
Wu, Y.; Wu, W.; Xing, C.; Zhou, M.; and Li, Z.
\newblock 2017.
\newblock Sequential matching network: A new architecture for multi-turn
  response selection in retrieval-based chatbots.
\newblock In {\em Proceedings of the 55th Annual Meeting of the Association for
  Computational Linguistics (Volume 1: Long Papers)}, volume~1,  496--505.

\bibitem[\protect\citeauthoryear{Yoshino \bgroup et al\mbox.\egroup
  }{2018}]{DSTC7}
Yoshino, K.; Hori, C.; Perez, J.; D'Haro, L.~F.; Polymenakos, L.; Gunasekara,
  C.; Lasecki, W.~S.; Kummerfeld, J.; Galley, M.; Brockett, C.; Gao, J.; Dolan,
  B.; Gao, S.; Marks, T.~K.; Parikh, D.; and Batra, D.
\newblock 2018.
\newblock The 7th dialog system technology challenge.
\newblock {\em arXiv preprint}.

\bibitem[\protect\citeauthoryear{Zhang \bgroup et al\mbox.\egroup
  }{2018}]{zhang2018modeling}
Zhang, Z.; Li, J.; Zhu, P.; Zhao, H.; and Liu, G.
\newblock 2018.
\newblock Modeling multi-turn conversation with deep utterance aggregation.
\newblock In {\em Proceedings of the 27th International Conference on
  Computational Linguistics},  3740--3752.

\bibitem[\protect\citeauthoryear{Zhou \bgroup et al\mbox.\egroup
  }{2016}]{zhou2016multi}
Zhou, X.; Dong, D.; Wu, H.; Zhao, S.; Yu, D.; Tian, H.; Liu, X.; and Yan, R.
\newblock 2016.
\newblock Multi-view response selection for human-computer conversation.
\newblock In {\em Proceedings of the 2016 Conference on Empirical Methods in
  Natural Language Processing},  372--381.

\bibitem[\protect\citeauthoryear{Zhou \bgroup et al\mbox.\egroup
  }{2018}]{zhou2018multi}
Zhou, X.; Li, L.; Dong, D.; Liu, Y.; Chen, Y.; Zhao, W.~X.; Yu, D.; and Wu, H.
\newblock 2018.
\newblock Multi-turn response selection for chatbots with deep attention
  matching network.
\newblock In {\em Proceedings of the 56th Annual Meeting of the Association for
  Computational Linguistics (Volume 1: Long Papers)}, volume~1,  1118--1127.

\end{thebibliography}
\bibliographystyle{aaai}

\end{document}